# NON-AUTOREGRESSIVE TRANSFORMER-BASED END-TO-END ASR USING BERT

*Fu-Hao Yu, and Kuan-Yu Chen*

National Taiwan University of Science and Technology, Taiwan

## ABSTRACT

Transformer-based models have led to significant innovation in classical and practical subjects as varied as speech processing, natural language processing, and computer vision. On top of the Transformer, attention-based end-to-end automatic speech recognition (ASR) models have recently become popular. Specifically, non-autoregressive modeling, which boasts fast inference and performance comparable to conventional autoregressive methods, is an emerging research topic. In the context of natural language processing, the bidirectional encoder representations from Transformers (BERT) model has received widespread attention, partially due to its ability to infer contextualized word representations and to enable superior performance for downstream tasks while needing only simple fine-tuning. Motivated by the success, we intend to view speech recognition as a downstream task of BERT, thus an ASR system is expected to be deduced by performing fine-tuning. Consequently, to not only inherit the advantages of non-autoregressive ASR models but also enjoy the benefits of a pre-trained language model (e.g., BERT), we propose a non-autoregressive Transformer-based end-to-end ASR model based on BERT. We conduct a series of experiments on the AISHELL-1 dataset that demonstrate competitive or superior results for the model when compared to state-of-the-art ASR systems.

## 1. INTRODUCTION

The central objective of an ASR system is to convert a given speech signal to its corresponding sequence of words. The importance of the technique stems from the fact that it is a key technology for humans to communicate with machines, a research topic that has attracted interest for several decades. Conventionally, ASR systems are usually composed of an acoustic model, a language model, a pronunciation lexicon, and a search algorithm. With the recent success of deep learning, much research has been devoted to assembling the various modules in an end-to-end manner, which not only eliminates the need for task-specific engineering but also prevents error propagation. Classic representations include the connectionist temporal classification (CTC) method [1], the RNN transducer model [2], the listen, attend, and spell (LAS) model [3], and the hybrid CTC/attention architecture [4], to name a few. Specifically, because of the reliability and capability of Transformer [5], more and more end-to-end ASR models are Transformer-based.

The spectrum of research on end-to-end ASR models can be categorized into autoregressive (AR) and non-autoregressive (NAR) models. Both employ an acoustic encoder to extract high-level representations from the speech signal; their major difference lies in the prediction philosophy to generate the corresponding word sequence for an input speech signal. AR methods predict each token conditioned on all of the previously generated tokens and the acoustic-level statistics distilled by the encoder. However, careful design can usually yield strong recognition results, but the modeling constitutes a decoding bottleneck. To mitigate the flaws of AR models, NAR models assume token-wise independence at the decoding stage to facilitate parallel implementations, usually resulting in fast inference and performance comparable to that of AR methods. Hence, NAR research has attracted much attention recently.

Deep learning is the focus of much recent research and many experiments in various applications because of its remarkable performance [6–8]. For the natural language processing (NLP) field, language representation methods can be viewed as pioneering studies [9, 10]. The line of research has lately involved manipulating large and deep neural network architectures typically composed of a stack of Transformers [11, 12–15, 16]. The major innovation of these models is capturing interactions between tokens via self-attention [5] and updating model parameters by optimizing the masked language model objective [17, 18], which modulates the constraint of the conventional left-to-right autoregressive objective [5]. One particularly successful approach is the bidirectional encoder representations from Transformers (BERT) model, whose success comes from a framework in which pre-training is followed by task-specific fine-tuning. Since BERT is trained on a set of extremely large-scale unsupervised data, it has learned general-purpose language knowledge. In the past years, BERT and variants such as XLNet [18] and RoBERTa [19] have dominated NLP.

On one hand, in view of the successes of BERT in various NLP-related tasks, we intend to reformulate speech recognition as a downstream task of BERT, thus an ASR system is expected to enjoy the benefits of a pre-trained language model and to be deduced by performing fine-tuning. On the other hand, to speed up the decoding process while retain competitive recognition performance, the paper strives to a novel non-autoregressive ASR model. Consequently, given the potentials of pre-trained language model and NAR ASR modeling, a non-autoregressive Transformer-based ASR model based on BERT (NAR-BERT-ASR) is proposed. Because of the NAR modeling, the model trains and decodes in parallel, making inference faster than in classic autoregressive modeling. Also, as NAR-BERT-ASR is coupled tightly with BERT, the capabilities and abilities of a pre-trained language model can be injected deeply into the ASR system. As a result, the NAR-BERT-ASR achieves competitive or superior results when compared to mature and/or state-of-the-art ASR models.

## 2. RELATED WORK

Generally, autoregressive (AR) ASR models decode each token conditioned on all of the previously generated tokens and the high-level acoustic representations distilled by the encoder. For a given speech utterance $O$, the objective of AR modeling is to generate an output token sequence $W$ by referring to $P(W|O)$. According to the chain rule, a common simplification is to decompose $P(W|O)$ into a series of conditional probabilities:

$$W^* = \underset{W}{\mathrm{argmax}}\, P(W|O) = P(w_1|O) \prod_{i=2}^{L} P(w_i|O, W_{<i}), \quad (1)$$

where $W_{<i} = \{w_1, \ldots, w_{i-1}\}$ denotes the partial token sequence before $w_i$. Although this design accounts for both acoustic- and text-level information, it makes AR ASR systems difficult to speed up decoding. In addition, ground-truth history tokens are fed to the decoder to predict the next token during training, while previous predictions from the decoder should be used as a condition to predict the upcoming token during inference. Thereby, scheduled sampling [20] is used to relax the mismatch between training and inference. Also note that the left-to-right nature of AR models limits the efficiency of parallel computation, making it difficult to speed up inference. Representative models include the RNN transducer model [2], the listen, attend, and spell (LAS) model [3], the speech Transformer [16], and the hybrid CTC/attention architecture model [4].

Orthogonal to AR models are non-autoregressive (NAR) ASR models, which assume that each token is conditionally independent of other tokens given a speech signal. That is, for a given speech utterance $O$, the NAR objective is

$$W^* = \underset{W}{\mathrm{argmax}}\, P(W|O) = \prod_{i=1}^{L} P(w_i|O). \quad (2)$$

Without the dependency between the current token and previous tokens, NAR models generate tokens in parallel to achieve a lower latency than AR models. Also, there is no training-inference mismatch in NAR models. Despite the apparent simplicity of NAR models in comparison to AR models, an increasing number of NAR-based ASR models have demonstrated fast inference and competitive performance compared to conventional AR methods. The listen and fill in missing letter (LFML) model [12], the listen attentively and spell once (LASO) model [13, 14], and the Mask CTC method [15] are all celebrated methods.

Transformer has been widely used in domains as varied as speech processing, natural language processing, and computer vision. The BERT model and its variants, all Transformer-based, constitute a new NLP paradigm. The self-supervised training objective ensures that these models can learn from a huge amount of unlabeled text data, and are thus capable of learning enriched language representations that are useful on a wide range of NLP tasks. To leverage the power of BERT for ASR, pioneer studies concentrate on transferring the knowledge from BERT in a teacher-student manner [14, 21] or using BERT as a strong language model to rescore candidate sentences [36, 37]. Empirical results, however, suggest only limited progress.

## 3. PROPOSED ASR MODEL

### 3.1. Architecture

In this section, we introduce NAR-BERT-ASR, a novel non-autoregressive Transformer-based end-to-end ASR model using BERT. As usual, the model consists of two components: an acoustic representation encoder and a text generation decoder. The acoustic extractor is inspired by the speech Transformer [16] and LASO [13, 14]; a pre-trained BERT model is placed on top of the encoder to be the decoder. Figure 1 depicts the architecture of the proposed ASR system.

Given a speech corpus containing $N$ training examples $\boldsymbol{D}_{ASR} = \langle O^{(j)}, W^{(j)}\rangle_{j=1}^{N}$, where the speech signal $O$ is quantized into a series of $T$ acoustic feature vectors $\{o_1, \ldots, o_T\}$ and $W$ denotes a sequence of $L$ tokens $\{w_1, \ldots, w_L\}$, we employ an acoustic encoder to produce high-level acoustic representations. Initially, for each speech signal, two convolutional neural network (CNN) layers with 2D filters and subsampling are introduced to capture short-term temporal variations and to refine the input acoustic feature frames. The stride of each CNN layer is set to 2, reducing the number of acoustic representations to a quarter of $T$. We add position embeddings to encode the absolute position of each acoustic representation within an utterance, and stack six Transformers to model long-term dependencies among all the acoustic statistics. After preliminary results, we made two adjustments in the Transformer implementation: we use gated linear units (GLUs) [22] instead of ReLU [14] as the activation function, and we use a pre-norm structure rather than the conventional post-norm structure to stabilize training [23]. The multi-head attention mechanism, layer normalization, residual dropout, and position-wise feedforward network are all unchanged [5]. After the Transformer sequence, we infer a set of high-level acoustic features $\{h_1^a, \ldots, h_{T/4}^a\} \in \mathbb{R}^{d_m \times (T/4)}$, where $d_m$ is a pre-defined hyperparameter.

As the time resolution of text and speech is at different scales, an important step of the proposed NAR-BERT-ASR is automatically determining the nonlinear alignment between an acoustic feature sequence and its corresponding text tokens. To make this idea work, a set of query vectors $\{q_1, \ldots, q_{L'}\} \in \mathbb{R}^{d_m \times L'}$ is initialized using sinusoidal functions [5]. The set of query vectors—actually the conventional position embeddings—and the high-level acoustic feature vectors $\{h_1^a, \ldots, h_{T/4}^a\}$ are fed into a Transformer, where the former is used to query the latter. The query vectors provide position-dependent information which can be interpreted as a set of ordered anchors; hence the process is designed to roughly organize the acoustic vectors and to generate $L'$ position-aware features. Subsequently, three Transformers, whose queries are output from the previous Transformer, while the keys and values are identical to $\{h_1^a, \ldots, h_{T/4}^a\}$, are used to iteratively refine the statistics to obtain a set of position-aware acoustic embeddings $\{h_1^p, \ldots, h_{L'}^p\} \in \mathbb{R}^{d_m \times L'}$. At the last stage of the acoustic encoder, the long-term dependencies among $\{h_1^p, \ldots, h_{L'}^p\}$ are again modeled by going through six Transformers and a simple feedforward network is then used to scale the vector dimension from $d_m$ to $d_n$, yielding a set of final acoustic embeddings $\{h_1^f, \ldots, h_{L'}^f\} \in \mathbb{R}^{d_n \times L'}$. In other words, the simple projection layer, which can have varying degrees of complexity, is design to bridge the dimension mismatch between the encoder and the decoder.

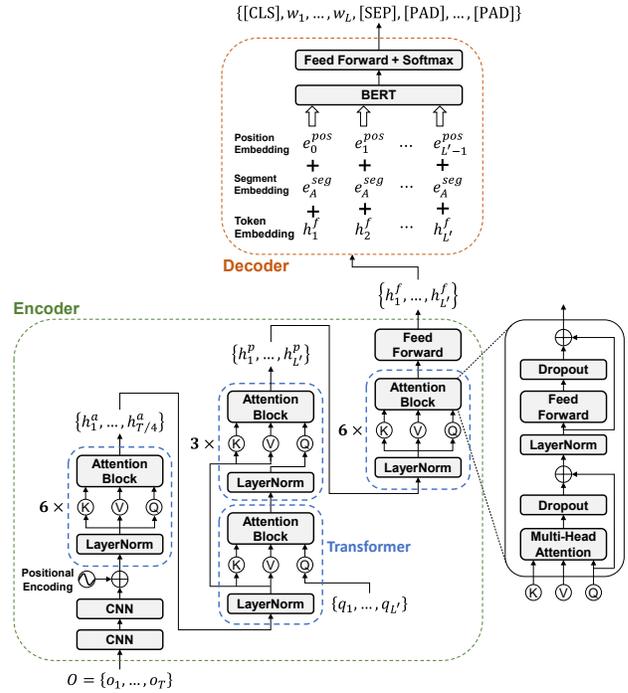

Figure 1: *Architecture of proposed non-autoregressive Transformer-based end-to-end ASR*

In order to harvest the power of BERT for ASR, we employ it as the decoder. As we believe that each final acoustic embedding $h_l^f$ carries specific phonetic information, we regard the inferred acoustic embeddings $\{h_1^f, \ldots, h_{L'}^f\}$ as token embeddings for BERT. Hence, a pre-trained BERT model is stacked upon the set of acoustic embeddings as a decoder. For the BERT-based decoder, the input vector sequence is a series of composite vectors that are sums of token embeddings, i.e., $\{h_1^f, \ldots, h_{L'}^f\}$, with their corresponding position embeddings and segment embeddings as in conventional BERT. A position-wise feedforward network with softmax activation is positioned on top of the BERT model. Consequently, speech recognition is framed as position-wise classification in a non-autoregressive manner. It is worthwhile to note that the design of the ASR architecture provides a natural way to deploy a pre-trained language model for an ASR system.

### 3.2. Training Recipe and Settings

Since NAR-BERT-ASR consists of an acoustic encoder to be trained from scratch and a decoder initialized by a pre-trained BERT model, the training process starts from pretraining the acoustic encoder to balance the whole system. To ensure that the final acoustic embeddings (i.e., $\{h_1^f, \ldots, h_{L'}^f\}$) contain specific phonetic information and can be used as token embeddings for the decoder, we stack a position-wise feedforward network whose parameters are initialized by the token embedding layer from BERT on top of the acoustic encoder. Softmax activation is used to create a probability distribution over tokens. Except for the last layer (i.e., the position-wise feedforward network), the model parameters $\theta_{enc}$ of the acoustic encoder are trained to minimize the negative log-likelihood loss:

$$\mathcal{L}_{enc} = \frac{1}{NL'} \sum_{j=1}^{N} \sum_{i=1}^{L'} \log P\left(w_i^{(j)} \middle| O^{(j)}, \theta_{enc}\right). \quad (3)$$

After pre-training, we discard the position-wise feedforward network and the acoustic encoder obtains its own initialization. Likewise, the entire model parameters are all then fine-tuned end-to-end by minimizing the negative log-likelihood loss.

Because the pre-trained BERT model is employed as the decoder, we use the BERT lexicon, which is composed of wordpieces, for the proposed NAR-BERT-ASR. All the transcriptions in the training set are tokenized by the BERT tokenizer, and the special tokens [CLS] and [SEP] are added at the beginning and end of each ground-truth text. Note that the hyperparameter $L'$ can be decided by counting the lengths in the training set. However, because during training, $L'$ is always larger or equal to the length of ground-truth text, the [PAD] token is added at the end of the text to match the length.

At the inference stage, since speech recognition is reformulated as position-wise classification, the output sequence is easily generated by concatenating the tokens with highest probability at each position individually:

$$W^* = \underset{W}{\mathrm{argmax}} \prod_{i=1}^{L'} P(w_i|O). \quad (4)$$

Although one difficulty in the decoding process is the independent assumption between tokens, several models account for this by integrating an extra language model with a beam search algorithm. It should be emphasized that one important innovation of the proposed NAR-BERT-ASR is that the text-level relationships are well-modeled naturally owing to the BERT-based decoder, allowing for a simple inference process.

In sum, the proposed NAR-BERT-ASR not only inherits the advantages of non-autoregressive ASR modeling but also explicitly leverages the power of BERT. Consequently, it facilitates parallel implementation, and provides a systematic and theoretically sound way to inject a pre-trained language model into an ASR system.

Table 1: *Statistics of the AISHELL-1 corpus*

|  |  | Training Set | Dev. Set | Test Set |
|---|---|---|---|---|
| #Utterances |  | 120,098 | 14,326 | 7,176 |
| #Hours |  | 150 | 18 | 10 |
| #Speakers |  | 340 | 40 | 20 |
| Duration (Sec.) | Min. | 1.2 | 1.6 | 1.9 |
|  | Max. | 14.5 | 12.5 | 14.7 |
|  | Avg. | 4.5 | 4.5 | 5.0 |
| #Tokens/Sentence | Min. | 1.0 | 3.0 | 3.0 |
|  | Max. | 44.0 | 35.0 | 21.0 |
|  | Avg. | 14.4 | 14.3 | 14.0 |

## 4. EXPERIMENTS

### 4.1. Experimental Setup

The experiments in this study were conducted on the AISHELL-1 dataset [34], which contains 178 hours of Mandarin speech. Table 1 shows some basic statistics of the collection. All the experiments used 80-dimensional log Mel-filter bank features with 3-dimensional pitch features, computed with a 25ms window size and shifted every 10ms. For the acoustic encoder, each CNN layer was composed of 32 filters with a kernel size of 3, a stride step of 2, and the ReLU activation function. The hyperparameter $d_m$ was set to 256, and $d_n$ was set to 768 to ensure consistency with the pre-trained BERT model. For the Transformer, each multi-head attention had 8 heads, and the hidden size of the feedforward neural network was set to 2048. Although the average length of the training transcriptions was 14.4, in this study $L'$ was set to 60. For the decoder, we directly used bert-base-chinese[1], with a vocabulary size of 21128, from Huggingface's Transformers library [24]. The NAR-BERT-ASR model is trained on an NVIDIA Tesla V100 GPU and is implemented with PyTorch [39] and ESPnet toolkit [40]. The pre-trained BERT model consisted of 12 Transformers. For training, the number of training epochs for both the acoustic encoder pre-training and the whole-model fine-tuning was 130. Each training batch contained 100 seconds of speech data with their corresponding text utterances, and the gradient was accumulated 12 times for each update. For the optimizer we used Adam, and to determine the learning rate we used the Noam scheduler, with 12000 warm-up steps. The dropout rate was set to 0.1 to avoid overfitting, and the label smoothing factor was set to 0.1. The SpecAugment technique [25] was used for data augmentation. We averaged the parameters of the last 10 epochs as our final model.

---

[1] https://huggingface.co/bert-base-chinese

Table 2: *Character error rate (CER) on AISHELL-1 for NAR-BERT-ASR vs. state-of-the-art approaches. Real-time factor (RTF) is computed as the ratio of the total inference time to the total duration of the test set.*

| Model | Dev. | Test | LM | Params | RTF |
|---|---|---|---|---|---|
| *Traditional hybrid ASR* | | | | | |
| Kaldi nnet3 | - | 8.6 | ✓ | | |
| Kaldi chain | - | 7.5 | ✓ | | |
| *Autoregressive ASR* | | | | | |
| BERT-ASR [32] | 54.6 | 58.8 | | | |
| LAS [26] | - | 8.7 | ✓ | 156M | |
| SA-T [28] | 8.3 | 9.3 | ✓ | | |
| ESPnet (Transformer) | 6.0 | 6.7 | ✓ | 58M | 0.4200 |
| Unsupervised pre-training [30] | - | 6.7 | | | |
| SAN-M [29] | 5.7 | 6.5 | | | |
| CAT [31] | - | 6.3 | ✓ | | |
| *Non-autoregressive ASR* | | | | | |
| CTC [33] | 7.8 | 8.7 | | | |
| LFML [12] | 6.2 | 6.7 | ✓ | | |
| LASO [14] | 5.9 | 6.9 | | 105.8M | |
| LASO (our implementation) | 5.9 | 6.7 | | | 0.0033 |
| NAR-Transformer [33] | 5.6 | 6.3 | | | |
| LASO with BERT [14] | 5.3 | 6.1 | | 105.8M | |
| NAR-BERT-ASR | **4.6** | **5.1** | | 119.5M | 0.0057 |

### 4.2. Experimental Results

We classify current ASR systems into traditional hybrid systems, autoregressive (AR) models, and non-autoregressive (NAR) models. At first glance, as shown in Table 2, most AR and NAR models outperform traditional hybrid models. Comparing the AR models with the NAR methods, we observe that NAR methods have caught up with classic AR models in terms of efficiency and effectiveness, which demonstrates the potential of NAR modeling.

In the second set of experiments, we compared the proposed NAR-BERT-ASR with state-of-the-art systems, as summarized in Table 2. We offer several observations. First, NAR-BERT-ASR outperforms all of the systems discussed in this study, which suggests the feasibility of the proposed ASR model, which integrates a pre-trained BERT model with a carefully designed acoustic encoder. Second, obviously, NAR models usually do not incorporate a language model in decoding, because using a language model no longer makes the model end-to-end and almost practical language models are autoregressive. Orthogonally, language model is almost a standard component of AR models, since empirical results indicate that language regularities and constrains are important clues for ASR. However, the tradeoff of the reduced recognition error rate is an increase in computational overhead. If we all agree upon that language model can provide useful information for ASR, NAR-BERT-ASR creates a potential way to install a language model in an end-to-end non-autoregressive ASR system. Consequently, NAR-BERT-ASR not only can deliver fast inference, but also achieves superior results than some mature and/or state-of-the-art ASR models.

Third, note that LASO can be seen as an important baseline, because it was the inspiration for the acoustic encoder of NAR-BERT-ASR. For improved performance, LASO uses the teacher-student learning [14, 35] to transfer knowledge from BERT[2]; in Table 2 this is denoted as "LASO with BERT". From the results, LASO with BERT can give about 11% relative improvement than LASO. When compared with the proposed ASR model, NAR-BERT-ASR achieves over 26% and 16% relative improvements than LASO and LASO with BERT, respectively, on the test set. The experiments reveal that the pre-trained language model (i.e., BERT) can indeed promote the recognition results, and the performance gap between LASO with BERT and NAR-BERT-ASR is because the latter is designed to tightly couple with BERT. In other words, during inference, NAR-BERT-ASR still explicitly and directly leverages the benefits from a pre-trained language model, while LASO with BERT only implicitly uses language knowledge transferred from BERT. Besides, LASO with BERT needs to bear additional computation costs to compute auxiliary ground-truths and losses during training [14]. In a nutshell, NAR-BERT-ASR is a preferable vehicle for utilizing BERT in NAR ASR system.

---
[2] https://storage.googleapis.com/bert_models/
2018_11_03/chinese_L-12_H-768_A-12.zip

Fourth, although BERT-ASR, an ASR system obtained purely by finetuning a pre-trained BERT model[3], achieves a mediocre result and remains several challenges [32], it is the cornerstone of the proposed NAR-BERT-ASR model. Taking this idea further, we leverage a more sophisticated acoustic encoder to distill high-level acoustic representations and also introduce an attention mechanism to eliminate the need for alignment information. Finally, in terms of the decoding speed measured in RTF, NAR-BERT-ASR is at least 70 times faster than the classical autoregressive model (i.e., ESPnet (Transformer)). LASO achieves slightly faster inference speed than NAR-BERT-ASR, while the latter can deliver remarkable improvements than LASO and LASO with BERT. Major reason for the difference of RTFs is the model architecture, where LASO has 13 Transformers and NAR-BERT-ASR consists of 28 Transformers. Hence, NAR-BERT-ASR suffers from the computational barrier made by the self-attention mechanism, which is a major component of Transformer. One of the emergent challenges is to speed up the inference process, and we leave as our future work. Based on the experiments, NAR-BERT-ASR, the end result, is thus a systematic and theoretically sound way to inject a pre-trained language model into a non-autoregressive ASR system.

### 4.3. Ablation Studies and Discussions

In Table 3, we further analyze different configurations of NAR-BERT-ASR. First, the performance of NAR-BERT-ASR is boosted further by increasing the number of training epochs during whole-model fine-tuning, although the improvement is not significant. Second, we seek to better understand the effect pre-training has on performance for both the acoustic encoder and decoder. The results show a large performance degradation if the whole model—both encoder and decoder—is trained from scratch. Furthermore, we found no good way to build a successful model by randomly initializing the acoustic encoder while copying the model parameters of the decoder from a pre-trained BERT model ("Without acoustic encoder pre-training"). One possible explanation for the two results is large difference between the time resolution of text and that of speech. This is addressed with a carefully designed acoustic encoder, which is thus more difficult to train than the decoder; to put it simply, the acoustic encoder pre-training is an important step for training. Orthogonally, respectable performance is achieved when only the decoder is initialized randomly ("Without decoder pre-training"). In addition, an interesting observation is the results of NAR-BERT-ASR without decoder pre-training are almost equivalent to LASO (cf. Table 2). A possible reason may be the fact that they have similar model architectures, and the model parameters are also close. Therefore, without the pre-trained BERT model, NAR-BERT-ASR can only achieve comparable results to LASO. Besides, when compared the without decoder pre-training with the original setting, we can conclude that the pre-trained BERT model brings 21% relative improvements for the proposed NAR-BERT-ASR. It is worthy to note again that NAR-BERT-ASR makes an effective and efficient way to leverage the power of BERT for ASR, thus the experimental results can show attractive improvements than practical state-of-the-art ASR models.

---
[3] https://github.com/ymcui/Chinese-BERT-wwm

Table 3: *Character error rate (CER) on AISHELL-1 for different configurations of NAR-BERT-ASR*

| Configuration | Dev. | Test |
|---|---|---|
| Original settings | 4.6 | 5.1 |
| Finetuning: 140 epochs | 4.5 | 5.0 |
| Without acoustic encoder pre-training | n/a | |
| Without decoder pre-training | 5.9 | 6.5 |
| Entire model trained from scratch | 15.4 | 18.7 |

### 5. CONCLUSION

We propose NAR-BERT-ASR, a novel Transformer-based non-autoregressive end-to-end ASR model based on BERT, which not only enjoys the benefits of a pre-trained language model but also inherits the strengths of a NAR model. In the experiments, the proposed model achieves promising results compared to several state-of-the-art methods. In the future, we will continue improving the architecture and exploring different training objectives for NAR-BERT-ASR. We also plan to replace BERT with other pre-trained language models (e.g., RoBERTa, XLNet and Albert), and evaluate the proposed ASR model on other benchmark corpora.

### 6. ACKNOWLEDGMENTS

This work was supported by the Ministry of Science and Technology of Taiwan under Grant MOST 110-2636-E-011-003 (Young Scholar Fellowship Program). We thank National Center for High-performance Computing (NCHC) for providing computational and storage resources.